\documentclass{article}
\usepackage[numbers,sort&compress]{natbib}
\usepackage{subfigure}
\usepackage{tabularx}
\usepackage{times}
\usepackage{multirow}
\usepackage{amsmath,amsthm,amssymb,latexsym}
\usepackage{booktabs}
\usepackage{threeparttable}
\usepackage{epsf}
\usepackage{graphicx}
\usepackage[justification=centering]{caption}
\newcommand{\be}{\begin{equation}}
\newcommand{\ee}{\end{equation}}
\newcommand{\ba}{\begin{array}}
\newcommand{\ea}{\end{array}}
\newcommand{\bea}{\begin{eqnarray}}
\newcommand{\eea}{\end{eqnarray}}
\newcommand{\reff}[1]{(\ref{#1})}
\renewcommand{\proof}{\noindent {\bf Proof} \quad}
\newcommand{\eproof}{\hfill{$\quad \Box$}}

\newtheorem{thm}{Theorem}[section]

\newtheorem{alg}{Algorithm}[section]

\newtheorem{lem}{Lemma}[section]
\newtheorem{rem}{Remark}[section]

\title{Joint Representation Classification for Collective Face Recognition\thanks{The work is partially supported by
the Chinese grants NSFC11471159, NSFC61170151 and Natural Science Foundation of Jiangsu Province (BK20141409).}}

\author {Liping Wang\thanks{Department of Mathematics,  Nanjing University of
Aeronautics and Astronautics, Nanjing 210016, China.\ Email:
wlpmath@nuaa.edu.cn.}\quad\quad Songcan Chen\thanks{Department of Computer Science and Engineering, Nanjing University of Aeronautics and Astronautics, Nanjing, 210016, China.\ Email:
s.chen@nuaa.edu.cn.}}

\begin{document}

\maketitle

\begin{abstract}
Sparse representation based classification (SRC) is  popularly used in many applications such as face recognition, and implemented in two steps: representation coding and classification. For a given set of testing images, SRC codes every image over the base images as a sparse representation then classifies it to the class with the least representation error. This scheme utilizes an individual representation rather than the collective one to classify such a set of images, doing so obviously ignores the correlation among the given images. In this paper, a joint representation classification (JRC) for collective face recognition is proposed. JRC takes the correlation of multiple images as well as a single representation into account. Under the assumption that the given face images are generally related to each other,  JRC codes all the testing images over the base images simultaneously to facilitate recognition. To this end, the testing inputs are aligned into a matrix and the joint representation coding is formulated to a generalized $l_{2,q}-l_{2,p}$-minimization problem.   To uniformly solve the induced optimization problems for any $q\in[1,2]$ and $p\in (0,2]$, an iterative quadratic method (IQM) is developed.  IQM is proved to be a strict descent algorithm with convergence to the optimal solution. Moreover, a more practical IQM is proposed for large-scale case. Experimental results on three public databases show that the JRC with practical IQM no only saves much computational cost but also achieves better performance in collective face recognition than the state-of-the-arts.


\end{abstract}

\section{Introduction}

Recently, representation coding  based classification  and its variants have been developed for face image recognition (FR) \cite{wright1,zhanglei,wright2,gao,huang}. This schemes achieve a great success in FR and  boost the applications of image classification \cite{wright3,wright4}. Sparse representation based classification (SRC) \cite{wright1} is the most known one which directly uses the sparse code for classification and efficiently recognizes the class giving the most compact representation. The main idea can be summarized to two steps: 1) coding a testing sample as a linear combination of all the training samples, then 2) classifying the testing sample to the most compact one by evaluating coding errors. Typical SRC employs the following $l_1$-minimization as the sparse representation model,
\be\label{eqno1}
\min\limits_{x}\|x\|_1\quad \hbox{s.t.}\ \|y-Ax\|_2\leq \varepsilon,
\ee
where $A\in R^{m\times d}$ is the dictionary of coding atoms and $y\in R^m$ is a given observation.  $x\in R^d$ is the coding vector and $\varepsilon>0$ denotes a noisy level. SRC outputs the identity of $y$ as
\be\label{eqno2}
\hbox{identity}(y)=\arg\min\limits_{1\leq i\leq I}\{\|y-Ax^*_i\|_2\},
\ee
where $I$ denotes the number of classes and $x^*_i$ is the coding coefficient vector associated with class $i$. The experimental results reported in \cite{wright1} exhibit that SRC scheme achieves amazing performance.  But the authors of \cite{zhanglei} argued that SRC over emphasized the importance of $l_1$-norm sparsity but ignored the effect of collaborative representation. Consequently, a collaborative representation based classification  with regularized least square (CRC-RLS)  was  presented in \cite{zhanglei} for face recognition
\be\label{eqno3}
\min\limits_{x}\|x\|_2\quad \hbox{s.t.}\ \|y-Ax\|_2\leq \varepsilon.
\ee
Anyway, problem (\ref{eqno3}) is easier to solve than (\ref{eqno2}) for its smoothness. Models (\ref{eqno2}) and (\ref{eqno3}) can be considered as the least square problems with different regularizers,
\be\label{eqno4}
\min\limits_{x}\|y-Ax\|_2^2+\lambda\|x\|_1\quad \hbox{and}\quad \min\limits_{x}\|y-Ax\|_2^2+\lambda\|x\|_2^2.
\ee
Moreover, Wright et al. \cite{wright2} ever used variant $l_1-$norm to improve the coding fidelity of $y$ over $A$,
\be\label{eqno5}
\min\limits_{x}\|y-Ax\|_1+\lambda\|x\|_1.
\ee
Actually, the models (\ref{eqno2})-(\ref{eqno5}) can be uniformly included in the framework
\be\label{eqno6}
\min\limits_{x}\|y-Ax\|_q^q+\lambda\|x\|_p^p,\quad 1\leq q\leq 2, \ 0<p\leq 2.
\ee
In (\ref{eqno6}), the representation and regularization measurements are extended to be $\|\cdot\|_q (\ 1\leq q\leq 2)$ and $\|\cdot\|_p (\ 0<p\leq 1)$ respectively. This modification provides possibility to adaptively choose the most suitable model for different applications. Moreover, the computational experiences \cite{Chartrand1,Chartrand2,xu} have showed that fractional norm $l_p \ (0<p<1)$ exhibits sparser pattern than $l_1$-norm. The unified generalization formula (\ref{eqno6}) is expected to achieve better performance.  On the other hand, model \reff{eqno6} is a vector representation based framework which implies the following weaknesses.

$\bullet$ Model \reff{eqno6} uses coding vector to represent testing samples one by one. In many face recognition, a great of number of images for each known subject have been collected from video sequence or photo album. The face recognition has to be conducted with a set of probe images rather than a single one \cite{mmd}. In this case, representation coding  based classification  like model \reff{eqno6} can not efficiently work.

$\bullet$  Any testing sample is coded independently from each other in \reff{eqno6}. This approach takes no account of the correlation hidden in the image set.  The difference and similarity between multiple pictures are totally ignored. It is well known that the collective faces share some similar feature patterns, such as eye or month pixels is more powerful in discrimination than those of forehead or cheek.

$\bullet$ When $q,p$ in (\ref{eqno6}) take different values, the involved optimization problems have to be solved by different algorithms. For example, \reff{eqno1} is solved by $l_1-l_s$ solver \cite{l1ls} or alternative direction of multiplier method while \reff{eqno3} chooses the algorithm presented in \cite{zhanglei}.

To overcome the weaknesses in (\ref{eqno6}) and make sufficient use of collective relationship among the given set of images, we consider to jointly represent all the test samples simultaneously over the training sample base. Here we employ matrix instead of vector as the coding variable to evaluate the distribution of feature space. This idea induces a joint representation based classification (JRC) for collective face recognition and reduces it to a $l_{2,q}-l_{2,p}$-minimization. To solve the derived optimization problem, a unified algorithm is designed and its convergence behavior is also analyzed. Experiments on three public face datasets validate the improvement of JRC over the state-of-the-arts.

This paper is organized as follows. In the second section, a joint representation based classification (JRC) will be established. The third section is dedicated to a unified algorithm for solving the special optimization problem induced by JRC. Some computational details are considered in the fourth section and an improved practical algorithm is proposed. The experimental results are reported in the last section.

\section{Joint Representation Classification for Collective Face Recognition}
\subsection{Joint Representation Model}
Suppose that we have $I$ classes of subjects in the dataset. $A_i\in R^{m\times d_i}(1\leq i\leq I)$ denotes the $i$-th class, and each column of $A_i$ is a sample of class $i$. Hence all the training samples are aligned by $A=[A_1,A_2,\cdots,A_I]\in R^{m\times d}$, where $d=\sum\limits_{i=1}^Id_i$. Given a collection of query images $y_1, y_2, \cdots, y_n\in R^m$, model \reff{eqno6} codes each $y_j\ (1\leq j\leq n)$ by the training samples $A$ as
\be\label{eqno2.0}y_j\approx Ax_j,
\ee
where $x_j\in R^d$ is the coding vector associated with $y_j$. If $y_j$ is from the $i-$th class, then $A_i$ is the most compact representation dictionary and the optimal solution $x_j^*$ to \reff{eqno6} can be used for classification. Obviously, coding pattern \reff{eqno2.0} depends on the single test sample $y_j$ individually for classification but takes no account of the correlation with other samples ($y_l, l\neq j$). Even though different frontal faces take on different appearances, they share similar features such as two eyes and brows at the  upper face while nose and mouth at the lower.  Difference and similarity of multiple face pictures form a unitary feature of the given set of images which play an important role for collective face recognition.

Denote $Y=[y_1,y_2,\cdots, y_n]\in R^{m\times n}$ all the query images, we propose to jointly represent the image set simultaneously by
\be\label{eqno2.1}
Y\approx AX,
\ee
where $X=[x_1,x_2,\cdots, x_n]\in R^{d\times n}$ stands for the collective coding matrix. As far as the columns are concerned, system \reff{eqno2.1} is an easy consequence of \reff{eqno2.0}. To measure the fidelity of the joint coding system (\ref{eqno2.1}), we consider $X$ in another sense.
Let $A^i\in R^d$ and $Y^i\in R^n$ be the $i-$th ($i=1,2,\cdots, m$)  row vectors of matrix $A$ and $Y$ respectively, formula (\ref{eqno2.1}) is equivalent to
\be\label{eqno2.2}
X^T(A^i)^T\approx(Y^i)^T \quad \hbox{for} \quad i=1,2,\cdots, m.
\ee
It is noticed that $A$ and $Y$ array the sampled images column by column, hence their rows span the feature space. In feature extraction view, the collective coding matrix $X$ also  projects the training feature space to approximate the testing feature space.  Traditional least square regression aims to minimize the error
\be\label{eqno2.3}
\min\limits_X\sum\limits_{i=1}^m\|X^T(A^i)^T-(Y^i)^T\|_2^2\quad \hbox{or} \quad \min\limits_X\sum\limits_{i=1}^m\|A^iX-Y^i\|_2^2\ .
\ee
Actually (\ref{eqno2.3}) can be easily reformulated as
\be\label{eqno2.4}
\min\limits_X\sum\limits_{i=1}^m\|(AX-Y)^i\|_2^2\ ,
\ee
where $(AX-Y)^i$ is the $i-$th row vector of $AX-Y$. Especially when the number of column in $AX-Y$ is $1$, the formula (\ref{eqno2.4}) is reduced to the fidelity function of (\ref{eqno4}). Then we prefer  a uniform generalization of (\ref{eqno4}) and (\ref{eqno5}) in the sense
\be\label{eqno2.5}
\sum\limits_{i=1}^m\|(AX-Y)^i\|_2^q\ ,\quad (1\leq q\leq 2).
\ee
Under the assumption that joint representation and feature distribution share the similar pattern for all testing face images, we use the following regularization
\be\label{eqno2.6}
\sum\limits_{i=1}^d\|X^i\|_2^p\ ,\quad (0< p\leq 2),
\ee
where $X^i$ is the $i-$th row vector of $X$ for $i=1,2,\cdots,d$.
Combining (\ref{eqno2.5}) and (\ref{eqno2.6}), we present the joint representation model for classification as follows
\be\label{eqno2.7}
\min\limits_X\sum\limits_{i=1}^m\|(AX-Y)^i\|_2^q+\lambda\sum\limits_{i=1}^d\|X^i\|_2^p,\quad (1\leq q\leq 2, 0< p\leq 2).
\ee
When the number of column in $Y$ is $1$, model (\ref{eqno2.7}) is reduced to coding vector version (\ref{eqno6}). Compared with coding vector $x$, joint coding matrix $X$ unites sample representation with feature projection which somewhat reflects the integral structure of dataset.  Hence (\ref{eqno2.7}) is a general extension of (\ref{eqno3})-(\ref{eqno6}). To simplify the formulation, we introduce the mixed matrix norm $l_{2,p}\ (p>0)$ (taking $\|X\|_{2,p}$ for example)
\be\label{eqno2.8}
\|X\|_{2,p}=(\sum\limits_{i=1}^d\|X^i\|_2^p)^{\frac{1}{p}},\quad X\in R^{d\times n},
\ee
where $X^i$ denotes the $i-$th row of $X$. Then (\ref{eqno2.7}) is rewritten as
\be\label{eqno2.9}
\min\limits_X\|AX-Y\|_{2,q}^q+\lambda\|X\|_{2,p}^p,\quad (1\leq q\leq 2, 0< p\leq 2).
\ee
Especially when $p\in (0,1)$, $l_{2,p}$ is not a valid matrix norm because it does not satisfy the triangular inequality of matrix norm axioms. Meanwhile the involved fractional matrix norm based minimization (\ref{eqno2.9}) is neither convex nor Lipschitz continuous  which brings computational challenge. Designing an efficient algorithm for such $l_{2,q}-l_{2,p}$-minimizations is very important. It is also the most challenging task in this paper.

\subsection{Joint Representation Based Classification}

For fixed parameter $q$ and $p$, suppose that $X^*$ is a minimizer of optimization problem (\ref{eqno2.9}), that is
\be\label{eqno2.10}
X^*=\arg\min\limits_X\|AX-Y\|_{2,q}^q+\lambda\|X\|_{2,p}^p\ .
\ee
If $X^*$ is partitioned to $I$ blocks as follows
\be\label{eqno2.111}
X^*=\left[\ba{c}
X^*_1\\ \vdots\\ X^*_i \\ \vdots\\X^*_I
\ea
\right]\ ,
\ee where $X_i^*\in R^{d_i\times n} \ (1\leq i\leq I)$. Let $\hat{X}_i^*$ denote the coding matrix associated with class $i$, that is
\be\label{eqno2.1111}
\hat{X}_i^*=\left[\ba{c}
0\\ \vdots\\ X^*_i \\ \vdots\\0
\ea
\right],
\ee
then $A\hat{X}_i^*=A_iX_i^*\ (1\leq i\leq I)$. For each testing image $y_j\ (j=1,2,\cdots,n)$, we classify $y_j$ to the class with the most compact representation. By evaluating the error corresponding to each class
\be\label{eqno2.11}
\|(Y-A\hat{X}_i^*)_j\|_2\ ,\quad i=1,2,\cdots, I
\ee
we pick out the index outputting the least error.
The joint representation based classification for face recognition can be concluded as follows.
\begin{alg}\label{alg1}(JRC scheme for FR)
\begin{enumerate}
\item Start: Given $A\in R^{m\times d}$, $Y\in R^{m\times n}$ and select parameters $\lambda>0$, $q\in [1,2]$ and $p\in (0,2]$.
\item Solve $l_{2,q}-l_{2,p}$-minimization problem (\ref{eqno2.9}) for coding matrix $X^*$.
\item For $j=1:n$\\
$
\ba{l}
For\ i=1:I\\
e_i(y_j)=\|(Y-A_iX_i^*)_j\|_2\\
end
\ea
$\\
Identity $(y_j)=\arg\min\limits_{1\leq i\leq I}\{e_i(y_j)\}$\\
end
\end{enumerate}
\end{alg}

When $n=1$, observation $Y$ contains only a single testing sample and JRC is reduced to vector representation based classification. Further on, SRC, CRC-RLS and $l_1$-norm fidelity model \reff{eqno5} are the special cases of JRC when $q=2\ \& \ p=1$, $q=p=2$ and $q=p=1$ respectively. In short, the main contributions of JRC lie in:
\begin{enumerate}
\item JRC implements collective face representation simultaneously. This scheme is more economical and efficient in computational cost and CPU time.  Moreover, JRC can handle image set based face recognition which broadens the applications of vector representation based classifications.
\item Joint coding technique fuses the difference of each testing sample representation and the similarity hidden in the feature space of multiple face images. For example, when $0<p\leq 1$ all query image are jointly represented by the training samples with the similarly sparse feature distribution.
\item  In the next section, a uniform algorithm will be developed to solve the optimization problem \reff{eqno2.9} for any $q\in [1,2]$ and $p\in (0,2)$. The algorithm is strict decreasing until it converges to the optimal solution to problem \reff{eqno2.9}. To the best of our knowledge, it is an innovative approach to solve such a generalized $l_{2,q}-l_{2,p}$-minimization.
\end{enumerate}

It is worth to point out that the JRC scheme can be easily extended for the presence of pixel distortion, occlusion or high noise in test images. Modify \reff{eqno2.1} as
\be\label{eqno2.12}
Y=AX+E\ ,
\ee
where $E\in R^{m\times n}$ is an error matrix. The nonzero entries of $E$ locate the corruption or occlusion in $Y$.  Substitute $\hat{A}=[A,\ I]\in R^{m\times(d+m)}$ and $\hat{X}=\left[\ba{l}X\\ E\ea\right]\in R^{(d+m)\times n}$ for $A$ and $X$ respectively, a stable joint coding model can be formulated to
\be\label{eqno2.13}
\min\limits_{\hat{X}}\|\hat{A}\hat{X}-Y\|_{2,q}^q+\lambda\|\hat{X}\|_{2,p}^p,\quad (1\leq q\leq 2, 0< p\leq 2).
\ee
Once a solution $\hat{X}^*=\left[\ba{l}X^*\\ E^*\ea\right]$ to \reff{eqno2.13} is computed, setting $Y^*=Y-E^*$ recovers a clean image from corrupted subject. To identity the testing sample $y_j$, we slightly modify the error of $y_j$ with each subject $e_i(y_j)=\|(Y-E^*-A_iX_i^*)_j\|_2$. Thus a robust JRC is an easy consequence of Algorithm \ref{alg1}. The corresponding algorithm and theoretical analysis can be similarly demonstrated. This paper will not concentrate on this subject.

\section{An Iterative Quadratic Method for JRC}

Obviously, efficiently solving optimization problem (\ref{eqno2.9})  plays the most important role in scheme \ref{alg1}. The mentioned models (\ref{eqno1}), (\ref{eqno3}) and (\ref{eqno5}) are special cases of (\ref{eqno2.9}), the algorithms used in \cite{wright1,zhanglei,wright2} to solve those special problems can not be directly extended. Such generally mixed matrix norm based minimizations as (\ref{eqno2.9}) have been widely used in machine learning. Rakotomamonjy and his co-authors \cite{Rakotomamonjy} proposed to use the mixed matrix norm $l_{q,p}\ (1\leq q<2, 0<p\leq 1)$ in multi-kernel and multi-task learning. But the induced optimization problems in \cite{Rakotomamonjy} have to be solved separately by different algorithms with respect to $p=1$ and $0<p<1$. For grouped feature selection, Suvrit \cite{Suvrit} addressed a fast projection technique onto $l_{1,p}$-norm balls particularly for $p=2,\infty$. But the derived method in \cite{Suvrit} does not match model (\ref{eqno2.9}). Similar joint sparse representation has been used for robust multimodal biometrics recognition in \cite{Sumit}. The authors of \cite{Sumit} employed the traditional alternating direction method of multipliers to solve the involved optimization problem. Nie et al. \cite{nie2} applied $l_{2,0+}$-norm to semi-supervised robust dictionary learning, while the optimization algorithm has not displayed definite convergence analysis.

In this section, a unified method will be developed to solve the $l_{2,q}-l_{2,p}$-minimization (\ref{eqno2.9}) for any $1\leq q\leq 2$ and $0<p\leq 2$. Especially when $p\in (0,1)$, (\ref{eqno2.9}) is neither convex nor non-Lipschitz continuous which results in much computational difficulties. Motivated by the idea of algorithm in \cite{wang1} for solving $l_{2,p}\ (0<p\leq 1)$-based minimization, we design an iteratively quadratic algorithm for such $l_{2,q}-l_{2,p}$-minimization. Moreover, the convergence analysis will be uniformly demonstrated.

\subsection{An Iteratively Quadratic Method}

After simply transformation, the definition of $\|X\|_{2,p}^p$ (\ref{eqno2.8}) can be rewritten as
\be\label{eqno3.1}
\|X\|_{2,p}^p=Tr(X^THX),
\ee
where
\be H=\left\{\ba{ll}\label{eqno3.2}
\hbox{diag}\{\frac{1}{\|X^1\|_2^{2-p}}, \frac{1}{\|X^2\|_2^{2-p}}, \cdots, \frac{1}{\|X^d\|_2^{2-p}}\},&\ \  p\in (0,2);\\
I,&\ \  p=2,
\ea\right.\ee
and $Tr(\cdot)$ stands for trace operation. If denote
\be G=\left\{\ba{ll}\label{eqno3.3}
\hbox{diag}\{\frac{1}{\|(AX-Y)^1\|_2^{2-q}}, \frac{1}{\|(AX-Y)^2\|_2^{2-q}}, \cdots, \frac{1}{\|(AX-Y)^m\|_2^{2-q}}\},& q\in [1,2);\\
I,& q=2,
\ea\right.\ee
the objective function of (\ref{eqno2.9}) can be reformulated to
\be\label{eqno3.4}
J(X):=Tr((AX-Y)^TG(AX-Y))+\lambda Tr(X^THX).
\ee
Hence the $KKT$ point of unconstrained optimization problem (\ref{eqno2.9}) is also the stationary point of $J(X)$,
\be\label{eqno3.5}
\frac{\partial J(X)}{\partial X}=qA^TG(AX-Y)+\lambda pHX=0\ ,
\ee
solving (\ref{eqno2.9}) is reduced to find the solution to equations \reff{eqno3.5}. If $A^TGA+\lambda\frac{p}{q}H$ is invertible, equation (\ref{eqno3.5}) is equivalent to
\be\label{eqno3.6}
X=(A^TGA+\lambda\frac{p}{q}H)^{-1}A^TGY.
\ee

To find the iterative solution to system (\ref{eqno3.6}), let us consider a closely related  optimization problem
\be\label{eqno3.7}
\min\limits_{X}\hat{J}(X):=Tr((AX-Y)^TG(AX-Y))+\lambda\frac{p}{q}Tr(X^THX).
\ee
$\hat{J}(X)$ is almost equivalent to $J(X)$ in spite of a scaled factor $\frac{p}{q}$ in regularization parameter.
If an iterative approximate solution $X_k$ to (\ref{eqno3.7}) has been generated, $G_k$  and $H_k$ can be derived from $X_k$ as definitions (\ref{eqno3.2}, \ref{eqno3.3}). Then we can compute the next iterative matrix $X_{k+1}$ by solving the following subproblem
\be\label{eqno3.8}
\min\limits_{X}Tr((AX-Y)^TG_k(AX-Y))+\lambda\frac{p}{q}Tr(X^TH_kX).
\ee

Actually, (\ref{eqno3.8}) is a scaled quadratic approximation to $J(X)$ at the iterative point $X_k$. Let $M_k=A^TG_kA+\lambda\frac{p}{q} H_k$, since $G_k$ and $H_k$ are usually symmetric and positive definite, problem (\ref{eqno3.8}) is equivalent to the following quadratic optimization problem
\be\label{eqno3.9}
\min\limits_{X}Q_k(X):={\frac{1}{2}}Tr(X^TM_kX)-Tr(Y^TG_kAX).
\ee
The minimizer to $Q_k(X)$ is also the solution to the linear system
\be\label{eqno3.10}
M_kX=A^TG_kY.
\ee

Based on the analysis and equations (\ref{eqno3.1}-\ref{eqno3.10}), the mixed $l_{2,q}-l_{2,p}$ ($1\leq q\leq 2, 0<p\leq 2$) norm based optimization problem (\ref{eqno2.9}) can be iteratively solved by a sequence of quadratic approximate subproblems. Hence we name this approach {\it iterative quadratic method (IQM)}. It is concluded as follows.
\begin{alg}\label{alg2}(IQM for Solving Problem (\ref{eqno3.7}))
\begin{enumerate}
\item Start: Given $A\in R^{m\times d}$, $Y\in R^{m\times n}$ and select parameters $\lambda>0$, $q\in [1,2]$ and $p\in (0,2]$.
\item Set $k=1$ and initialize $X_1\in R^{d\times n}$.
\item For $k=1,2,\cdots$ until convergence do :\\
$
\ba{l}
H_{k}=\hbox{diag}\{\frac{1}{\|X_k^i\|_2^{2-p}}\}_{i=1}^d\ (0<p<2)\ or\ H_k=I_d\ (p=2);\\
C_k=-Y;\\
For\ i=1:I\\
\ba{l}
B_i=A_i(X_k)_i;\\
C_k=B_i+C_k;
\ea\\
end\\
G_{k}=\hbox{diag}\{\frac{1}{\|C_k^i\|_2^{2-q}}\}_{i=1}^m\ (1\leq q<2)\ or\ G_k=I_m\ (q=2);\\
M_k=A^TG_kA+\lambda\frac{p}{q} H_k;\\
X_{k+1}=M_k^{-1}A^TG_kY.
\ea
$
\end{enumerate}
\end{alg}

It is noticed that each iteration has to compute the inverse of $M_k$ in Algorithm \ref{alg2} which is expensive and unstable. Here we suggest to employ the general Penrose inverse of $M_k$ to update the $X_{k+1}$. Moreover,  the main computation $A_iX^*_i$ for classification is a by-product of $B_i$ in computing the approximate solution  $X^*$. Hence identifying test images can be achieved with minor extra calculations.

Algorithm \ref{alg2} is a unified method solving $l_{2,q}-l_{2,p}-$minimizations for $q\in[1,2]$ and $p\in(0,2]$. This approach provides algorithmic support to adaptively choose better fidelity measurement and regularization in various applications. Especially IQM provides a uniform algorithm for solving the existed representation based models: sparse representation ($q=2,\ p=1$), collaborative representation ($q=p=2$) and $l_1$-norm face recognition ($q=p=1$).

\subsection{Convergence Analysis of IQM}

In this part, we will demonstrate the theoretical convergence of Algorithm \ref{alg2}.  The key point is that the objective function $J(X)$ strictly  decreases  with respect to iterations until the matrix sequence $\{X_k\}$ converges to a stationary point of $J(X)$.

\begin{lem}\label{lem1}
Let $\varphi(t)=t-at^{\frac{1}{a}}$, where $a\in (0, 1)$. Then for any $t>0$, $\varphi(t)\leq 1-a$, and  $t=1$ is the unique maximizer.
\end{lem}

\proof Taking the derivative of $\varphi(t)$ and set to zero, that is
$$\varphi^\prime(t)=1-t^{\frac{1}{a}-1}=0\ ,
$$ then  $\varphi^\prime(t)=0$ has the unique solution $t=1$ for any $a\in (0, 1)$ which is just the maximizer of $\varphi(t)$ in $(0, +\infty)$.\eproof

\begin{lem}\label{lem2}
Given $X_k$ and $X_{k+1}$ in $R^{d\times n}$,  the following inequalities hold,
\be\label{eqno3.11}
\|AX_{k+1}-Y\|_{2,q}^q-\frac{q}{2}\sum\limits_{i=1}^m\frac{\|(AX_{k+1}-Y)^i\|_2^2}{\|(AX_{k}-Y)^i\|_2^{2-q}}\leq (1-\frac{q}{2})\|AX_k-Y\|_{2,q}^q
\ee
and
\be\label{eqno3.12}
\|X_{k+1}\|_{2,p}^p-\frac{p}{2}\sum\limits_{i=1}^d\frac{\|X_{k+1}^i\|_2^2}{\|X_{k}^i\|_2^{2-p}}\leq (1-\frac{p}{2})\|X_k\|_{2,p}^p
\ee  for any $q \in [1,2)$ and $p \in (0,2)$.
Moreover, the equalities in Eq. (\ref{eqno3.11}) and (\ref{eqno3.12}) hold if and only if $\|(AX_{k+1}-Y)^i\|_2=\|(AX_{k}-Y)^i\|_2$ for $i=1,2,\cdots,m$ and $\|X_{k+1}^i\|_2=\|X_{k}^i\|_2$ for $i=1,2,\cdots,d$.
\end{lem}

\proof Substituting $t_1=\frac{\|(AX_{k+1}-Y)^i\|_2^q}{\|(AX_k-Y)^i\|_2^q}$ and setting $a_1=\frac{q}{2}$ in Lemma \ref{lem1}, we obtain
\be\label{eqno3.13}
\frac{\|(AX_{k+1}-Y)^i\|_2^q}{\|(AX_k-Y)^i\|_2^q}-\frac{q}{2}\frac{\|(AX_{k+1}-Y)^i\|_2^2}{\|(AX_k-Y)^i\|_2^2}\leq 1-\frac{q}{2}\ .
\ee Similarly taking $t_2=\frac{\|X_{k+1}^i\|_2^p}{\|X_{k}^i\|_2^p}$ and $a_2=\frac{p}{2}$ in $\varphi(t)$, we have
\be\label{eqno3.14}
\frac{\|X_{k+1}^i\|_2^p}{\|X_{k}^i\|_2^p}-\frac{p}{2}\frac{\|X_{k+1}^i\|_2^2}{\|X_k^i\|_2^2}\leq 1-\frac{p}{2}\ .
\ee
Multiplying Eq. (\ref{eqno3.13}) and Eq. (\ref{eqno3.14}) by $\|(AX_k-Y)^i\|_2^q$ and $\|X_k^i\|_2^p$ respectively, we have the following inequalities simultaneously
\be\label{eqno333}
\|(AX_{k+1}-Y)^i\|_2^p-\frac{q}{2}\frac{\|(AX_{k+1}-Y)^i\|_2^2}{\|(AX_{k}-Y)^i\|_2^{2-q}}\leq (1-\frac{q}{2})\|(AX_k-Y)^i\|_2^q
\ee
for $i=1,2,\cdots, m$,  and
\be\label{eqno444}
\|X_{k+1}^i\|_2^p-\frac{p}{2}\frac{\|X_{k+1}^i\|_2^2}{\|X_{k}^i\|_2^{2-p}}\leq (1-\frac{p}{2})\|X_k^i\|_2^p,\quad\ i=1,2,\cdots, d\ .
\ee
Summing up $i$ in formulas $(\ref{eqno333})$ and $(\ref{eqno444})$, we can derive $(\ref{eqno3.11})$ and $(\ref{eqno3.12})$.

Based on Lemma \ref{lem1},  $t_1=1$ and $t_2=1$ are the unique minimizers for $\varphi(t)$ in $(0,+\infty)$ when $a_1=\frac{q}{2}$ and $a_2=\frac{p}{2}$ respectively. Namely,  $\|(AX_{k+1}-Y)^i\|_2=\|(AX_{k}-Y)^i\|_2$ and  $\|X_{k+1}^i\|_2=\|X_{k}^i\|_2$ are necessary and sufficient for equalities hold in (\ref{eqno333}) and (\ref{eqno444}) respectively. \eproof

\begin{rem}\label{rem1}
(\ref{eqno3.11}) and (\ref{eqno3.12}) are established nothing to do with Algorithm \ref{alg2}. The inequalities express the innate properties of mixed matrix norms $l_{2,q}-l_{2,p}$ for $q\in [1,2)$ and $p\in (0,2)$.
\end{rem}

\begin{thm}\label{thm1}
Suppose that $\{X_{k}\}$ is the matrix sequence generated by Algorithm \ref{alg2}. Then $J(X_k)$  strictly decreases with respect to  $k$ for any $1\leq q\leq 2$ and $0<p\leq 2$ until $\{X_{k}\}$ converges to a stationary point of $J(X)$.
\end{thm}

\proof Based on the procedure of Algorithm \ref{alg2}, $X_{k+1}$ is the solution to linear system (\ref{eqno3.10}), also the optimal matrix of problems (\ref{eqno3.8}) and (\ref{eqno3.9}). Thus we have
\be\label{eqno3.15}
Q_k(X_{k+1})\leq Q_k(X_k)\ .
\ee
For $q\in [1,2)$ and $p\in (0,2)$, (\ref{eqno3.15}) is equivalent to
\be\label{eqno3.16}
q\sum\limits_{i=1}^m\frac{\|(AX_{k+1}-Y)^i\|_2^2}{\|(AX_{k}-Y)^i\|_2^{2-q}}+\lambda p\sum\limits_{i=1}^d\frac{\|X_{k+1}^i\|_2^2}{\|X_{k}^i\|_2^{2-p}}
\leq q\|AX_k-Y\|_{2,q}^q+\lambda p\|X_k\|_{2,p}^p\ ,
\ee
It is noticed that $J(X_k)=\|AX_k-Y\|_{2,p}^p+\lambda\|X_k\|_{2,p}^p$. Adding inequalities (\ref{eqno3.11}) and  $\lambda\cdot$(\ref{eqno3.12}), the following formula will be derived
\be\label{eqno3.17}\ba{l}
J(X_{k+1})-(\frac{q}{2}\sum\limits_{i=1}^m\frac{\|(AX_{k+1}-Y)^i\|_2^2}{\|(AX_{k}-Y)^i\|_2^{2-q}}+
\lambda\frac{p}{2}\sum\limits_{i=1}^d\frac{\|X_{k+1}^i\|_2^2}{\|X_{k}^i\|_2^{2-p}})\\ \leq J(X_k)-(\frac{q}{2}\|AX_k-Y\|_{2,q}^q+\lambda\frac{p}{2}\|X_k\|_{2,p}^p )\ .
\ea\ee
Based on (\ref{eqno3.16}) and (\ref{eqno3.17}), $J(X_{k+1})\leq J(X_k)$ can be easily derived for $q\in [1,2)$ and $p\in (0,2)$.

For $q=2$ or $p=2$, the inequalities is much easier to derive. Taking $q=2$ and $p\in (0,2)$ for example, (\ref{eqno3.15}) is reduced to
\be\label{eqno66}
\|AX_{k+1}-Y)\|_{2,2}^2+\lambda \frac{p}{2}\sum\limits_{i=1}^d\frac{\|X_{k+1}^i\|_2^2}{\|X_{k}^i\|_2^{2-p}}
\leq \|AX_k-Y\|_{2,2}^2+\lambda\frac{p}{2}\|X_k\|_{2,p}^p\ ,
\ee Combining the formulas (\ref{eqno66}) and (\ref{eqno3.12}), we also obtain $J(X_{k+1})\leq J(X_k)$. In the case of $q=2,\ p\in (0,2)$ or $q=p=2$, $J(X_{k+1})\leq J(X_k)$ can be deduced analogously.

Once $J(X_{k+1})=J(X_k)$ happens for some $k$, the equalities in (\ref{eqno3.16}) and (\ref{eqno3.17}) (or (\ref{eqno66})) hold. Hence the equalities in (\ref{eqno3.11}) and (\ref{eqno3.12}) are active. From Lemma \ref{lem2}, we obtain  $\|(AX_{k+1}-Y)^i\|_2=\|(AX_{k}-Y)^i\|_2$ for $i=1,2,\cdots,m$ and $\|X_{k+1}^i\|_2=\|X_{k}^i\|_2$ for $i=1,2,\cdots,d$. Thus $G_{k+1}=G_k$ and $H_{k+1}=H_k$ which implies that $X_{k+1}$ is a solution to (\ref{eqno3.6}). \eproof

The objective function sequence $\{J(X_k)\}$ is decreasing and lower bounded. Hence $\{J(X_k)\}$ eventually converges to some minimum of problem \reff{eqno2.9}. The descending quantity measures the convergence precision.

\begin{rem}\label{rem2}
The stopping criterion of Algorithm \ref{alg2} can be chosen as $J(X_k)-J(X_{k+1})\leq \epsilon$ or $\rho_k:=\frac{J(X_k)-J(X_{k+1})}{J(X_k)}\leq \epsilon$ for some required precision  $\epsilon>0$.
\end{rem}

Theoretically, $X_k^i=0$ or $C_k^i=0$ likely occurs in some step $k$, then $H_k$ and $G_k$ can not be well updated for non-Frobenius norm case ($0<p<2$ and $1\leq q<2$). We deal with it by perturbing with $\delta>0$ such that $\{H_k\}_{ii}=\delta^{p-2}> 0$ and $\{G_k\}_{ii}=\delta^{q-2}> 0$. The descending of $\{J(X_k)\}$ is relaxed to
\be\label{eqno67}
J(X_{k+1})\leq J(X_k)+(1-\frac{p}{2})\delta^p\quad \hbox{or}\quad J(X_{k+1})\leq J(X_k)+(1-\frac{q}{2})\delta^q\ .
\ee
If the convergence precision  $\epsilon$ is chosen fairly larger than perturbation $\delta$ ($\epsilon\gg \delta$), perturbed $J(X_k)$ can be still considered approximate decreasing. As a matter of fact, $X_k^i=0$ and $C_k^i=0$ never happen in practical implementation.

\section{Practical Implementation of JRC}

In Algorithm \ref{alg2}, IQM has to update the matrix sequence by computing the inverse matrix of $M_k$. It is expensive in practical implementation especially for large scale problems. Reviewing the procedure of Algorithm \ref{alg2}, we notice that $X_{k+1}=M_k^{-1}A^TG_kY$ exactly solves the $k-$th subproblem (\ref{eqno3.9}) which is unnecessary. It is observed that (\ref{eqno3.9}) is a quadratic positive definite subproblem. There are a lot of efficient algorithms to solve it approximately, such as conjugate gradient method, gradient methods with different stepsizes, etc. In this paper, we choose Barzilai and Borwein (BB) gradient method due to its simplicity and efficiency. BB gradient method was firstly presented in \cite{bb}, afterwards extended and developed in many occasions and applications \cite{bb,raydan1, raydan2, dai1, dai2, yuan}. When applied to quadratic matrix optimization subproblem (\ref{eqno3.9}), the  Barzilai and Borwein gradient method takes on
\be\label{eqno4.1}
X_k^{(t+1)}=X_k^{(t)}-\alpha_k^{(t)}\nabla Q_k(X_k^{(t)}),
\ee
where the superscript $(t)$  denotes the $t-$th iteration solving \reff{eqno3.9}. $\nabla Q_k(X_k^{(t)})$ is the gradient matrix of $Q_k(X)$ with respect to $X_k^{(t)}$
\be\label{eqno4.2}
\nabla Q_k(X_k^{(t)})=M_kX_k^{(t)}-A^TG_kY\ .
\ee
The  Barzilai and Borwein gradient method \cite{bb} chose the stepsize $\alpha_k^{(t)}$ such that $D_k^{(t)}=\alpha_k^{(t)}I$ has a certain quasi-Newton property
\be\label{eqno4.3}
D_k^{(t)}=\arg\min\limits_{D=\alpha I}\|S_{k}^{(t-1)}-DT_k^{(t-1)}\|_F
\ee
or
\be\label{eqno4.3'}
D_k^{(t)}=\arg\min\limits_{D=\alpha I}\|D^{-1}S_{k}^{(t-1)}-T_k^{(t-1)}\|_F,
\ee
where $\|\cdot\|_F$ denotes Frobenius matrix norm and $S_{k}^{(t-1)},\ T_{k}^{(t-1)}$ are determined by the information achieved at the points $X_k^{(t)}$ and $X_k^{(t-1)}$
\be\label{eqno4.4}
\ba{l}
S_{k}^{(t-1)}:=X_k^{(t)}-X_k^{(t-1)}\ ;\\
T_{k}^{(t-1)}:=\nabla Q_k(X_k^{(t)})-\nabla Q_k(X_k^{(t-1)})=M_kS_{k}^{(t-1)}\ .
\ea\ee
Solving (\ref{eqno4.3}) yields two BB stepsizes
\be\label{eqno4.5}
\alpha_k^{(t)}={\frac{Tr((S_k^{(t-1)})^TT_k^{(t-1)})}{Tr((T_k^{(t-1)})^TT_k^{(t-1)})}}
\ee and
\be\label{eqno4.5'}
\alpha_k^{(t)}={\frac{Tr((S_k^{(t-1)})^TS_k^{(t-1)})}{Tr((S_k^{(t-1)})^TM_kS_k^{(t-1)})}}\ .
\ee

Compared with the classical steepest descent method, BB gradient method often needs less computations but converges more rapidly \cite{fletcher}. For  optimization problems higher than two dimensions, BB method has theoretical difficulties due to its heavy non-monotone behavior. But for strongly convex quadratic problem with any dimension, BB method is convergent at $R-$linear rate \cite{raydan1,dai1}. BB method has also been applied to matrix optimization problem \cite{jiang} and exhibited desirable performance. Based on equations (\ref{eqno4.1})-(\ref{eqno4.5'}), the last step in Algorithm \ref{alg2}, $X_{k+1}=M_k^{-1}A^TG_kY$, can be  practically substituted by the BB gradient method as the $k-$th inner loop.
\begin{alg}\label{alg3}(BB Gradient Method for Solving Subproblem (\ref{eqno3.9}))
\begin{enumerate}
\item Start: given the inner loop stopping criterion $\epsilon_2>0$
\item Initialize $X_k^{(1)}=X_k$ and $\nabla Q_k^{(1)}=M_kX_k^{(1)}-A^TG_kY$;\\
\item For $t=1,2,\cdots$ until $Tr(\nabla Q_k^{(t)})\leq \epsilon_2$, output $X_{k+1}=X_k^{(t)}$, do :\\
$
\ba{l}
\hbox{if}\quad t=1\\
\ba{l}
\alpha_k^{(t)}={\frac{Tr((\nabla Q_k^{(t)})^T\nabla Q_k^{(t)})}{Tr((\nabla Q_k^{(t)})^TM_k\nabla Q_k^{(t)})}};\\
\ea\\
\hbox{else}\\
\ba{l}
S_{k}^{(t-1)}=X_k^{(t)}-X_k^{(t-1)};\\
T_{k}^{(t-1)}=\nabla Q_k^{(t)}-\nabla Q_k^{(t-1)};\\
\alpha_k^{(t)}\quad\hbox{is computed as}\quad (\ref{eqno4.5})\quad\hbox{or}\quad\reff{eqno4.5'};\\
\ea\\
\hbox{end}
\ea\\
X_k^{(t+1)}=X_k^{(t)}-\alpha_k^{(t)}\nabla Q_k^{(t)};\\
\nabla Q_k^{(t+1)}=M_kX_k^{(t+1)}-A^TG_kY;\\
$
\end{enumerate}
\end{alg}

In the $k-$th inner loop, Algorithm \ref{alg3} chooses two initial matrices. One is the approximate solution $X_k$ to the last subproblem and another one is the Cauchy point from  $X_k$ \cite{Cauchy} . The Cauchy stepsize $\alpha_k^{(1)}$ is the solution to the one-dimensional optimization problem
\be\label{eqno4.6}
\min\limits_{\alpha>0}\phi(\alpha):=Q_k(X_k-\alpha\nabla Q_k(X_k))\ ,
\ee then the Cauchy point is $X_k+\alpha_k^{(1)}\nabla Q_k(X_k))$.
If $M_k$ in Algorithm \ref{alg2} is guaranteed to be positive definite (if not, $H_k$ or $G_k$ can be slightly perturbed), subproblem (\ref{eqno3.9}) is a strongly convex quadratic.  BB gradient method with step length (\ref{eqno4.5}) or (\ref{eqno4.5'}) will  converges at $R-$linear rate.

For simplicity, we name the IQM with inexact Algorithm \ref{alg3} practically iterative quadratic method (PIQM). Still denote $\{X_k\}$ the approximate matrix sequence generated by PIQM. BB inner loop makes the objective function value of subproblem (\ref{eqno3.9}) decline, that is $Q(X_{k+1})\leq Q(X_k)$. Then $\{J(X_k)\}$ is always decreasing which is sufficient and necessary for $\{X_k\}$ uniformly converging to the stationary point of problem (\ref{eqno2.9}). The following conclusion can be easily derived.
\begin{thm}\label{thm3}
Denotes $X^*$ the output point generated by PIQM, then $X^*$ is an approximate stationary point of $J(X)$. Especially for $q,\ p\in [1,2]$,  $X^*$ is an approximate global  minimizer of optimization problem (\ref{eqno2.9}). When $p$ is fractional,  $X^*$ is  one of KKT points.
\end{thm}

An practical version of iteratively quadratic method for joint classification in face recognition can be concluded as follows.
\begin{alg}\label{alg4}(PIQM for JRC)
\begin{enumerate}
\item Start: loading $A, Y$ and setting $\lambda>0$, $q\in [1,2], p\in (0,2]$ and precision levels $\varepsilon_1>0,\varepsilon_2>0$.
\item Employing PIQM to solve \reff{eqno2.9}, output an approximate coding matrix $X^*:=X_{k+1}$.
\item Classifying $Y$ by $X^*$.
\end{enumerate}
\end{alg}

\section{Experimental Results}

In this section, the joint representation based classification (JRC) with PIQM will be applied to face recognition. Three public data sets are used. Brief description is given as follows.
\begin{description}
\item[AT\&T database] is formerly known ``the ORL database of faces". It consists of $400$ frontal images for $40$ individuals. For each suject, $10$ pictures were taken at different times, with varying lighting conditions, multiple facial expression, adornments and rotations up to $20$ degree. All the images are aligned with dimension $112\times 92$. The database can be retrieved from $http://www.cl.cam.ac.uk/\\ Research/DTG/attarchive:pub/data/att_faces.tar.Z$ as a 4.5Mbyte compressed tar file. Typical pictures can be seen in Figure \ref{fig1}.
\begin{figure}[!htb]
\centering
\centering
\includegraphics[width=90mm]{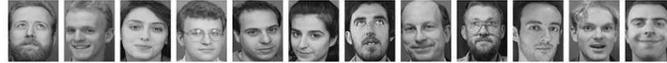}
\caption{\label{fig1} Typical images of AT \& T database}
\end{figure}
\item[Georgia-Tech database] contains $15$ images each of $50$ subjects. The images are taken in two or three sessions at different times with different facial expressions, scale and background. The average size of the faces in these images is $150\times 150$ pixels. Georgia Tech face database and the annotation can be found in \\ $http://www.anefian.com/research/face_reco.htm$. Typical pictures of four persons are shown in Figure \ref{fig2}.
\begin{figure}[!htb]
\centering
\centering
\includegraphics[width=100mm]{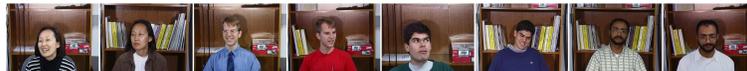}
\caption{\label{fig2}Typical images of Georgia-Tech database}
\end{figure}
\item[Extended Yale B database] consists of $2414$ frontal-face images of $38$ subjects. Each subject has around $64$ images. The images are cropped and normalized to $192\times 168$ under various laboratory-controlled lighting conditions  \cite{yaleB1,yaleB2}. Figure \ref{fig3} displays typical pictures of 4 subjects.
\begin{figure}[!htb]
\centering
\includegraphics[width=90mm]{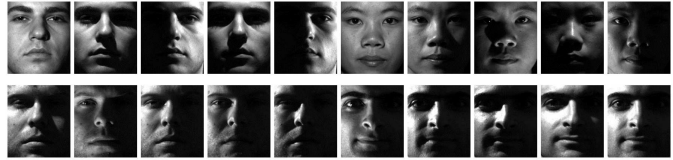}
\caption{\label{fig3} Typical images of Extended Yale B database}
\end{figure}
\end{description}

Extensive experiments are conducted for different image sizes and different parameters. Four comparable schemes are implemented, JRC, SRC, CRC-RLS and traditional SVM classifier. JRC is practically carried out via PIQM while SRC is solved by $l_1-l_s$ solver \cite{l1ls} and CRC-RLS employs the code from \cite{zhanglei}. We realize SVM by the software LIBSVM \cite{libsvm} with linear kernel, the pseudo code can be found in $http://www.csie.ntu.edu.tw/~cjlin/libsvm/faq.html\sharp f203$. All the schemes are implemented by Matlab R2014a(win32) on a typical 4GiB memory and 2.40GHz PC. 		

Considering that JRC is a joint framework including SRC and CRC-RLS, we select six pairs of $q,\ p$ in $[1,2]$ and $(0,2]$ respectively:\\
$$\ba{l}
q=p=2\ (\hbox{corresponding to CRC-RLS}), \\
q=2,\ \ p=1\ (\hbox{corresponding to SRC}),
\ea $$\\
and other four generalized cases \\
$$\ba{ll}
q=1.5\ \& \ p=1, & q=1.5 \ \& \ p=0.5,\\
q=1\ \& \ p=1, & q=1 \  \& \ p=0.5.
\ea
$$
The parameter $\lambda$ in \reff{eqno2.9} is varied from $0.01$ to $10$ each $10$ times, and the best result is picked out. All the stopping precisions are set $10^{-3}$.

All the images are re-sized like that of \cite{wright1, zhanglei}. For AT\&T database, the pictures are down sampled to $11\times 10$. The downsampling ratios of  Georgia-Tech database and Extended Yale B database are $1/8$ and $1/16$. For each subject, around $80\%$ pictures are randomly selected for training and the left for testing. For example, $8$ pictures of each individual in AT\&T database are randomly picked out for training while the left $2$ are for testing. All the classification schemes are directly applied to the images without any pre-processing. The recognition accuracy and running time are reported in Table \ref{tab1}-\ref{tab3}.
\begin{table*}[!htb]
\centering
 \begin{tabular}{lccc}
 \hline
 Methods & The recognition accuracy &  & CPU time\\
\hline\hline
SRC  &98.75	&    &  67.2658 \\
JRC(q=2,p=1) & 97.5	&  & 0.1612  \\
CRC-RLS   & 95	&    & 0.0872 \\
JRC(q=p=2) & 97.5	&    &  0.0073\\
SVM & 95 &   &  0.0667\\
JRC(q=1.5,p=1)& 97.5 & & 0.3867 \\
JRC(q=1.5,p=0.5)&95 & & 1.8756 \\
JRC(q=p=1)& 97.5 && 0.1994 \\
JRC(q=1,p=0.5)& 97.5 && 0.1640\\
\hline
\end{tabular}
\caption{\label{tab1} The recognition accuracy (\%) and running time (second)\protect \\ for AT\&T database}
\end{table*}

\begin{table*}[!htb]
\centering
 \begin{tabular}{lccccc}
 \hline
 & \multicolumn{2}{c}{Downsampling ratio 1/8}& &\multicolumn{2}{c}{Downsampling ratio 1/16}\\
\hline
Methods  &  Accuracy    & Time & &  Accuracy  & Time \\ \hline\hline	
SRC      &  99.33	    & 2843 & &	97.33     & 3197 \\
JRC(q=2,p=1) & 99.33	& 2.41 & &  97.33     & 1.07 \\
CRC-RLS & 98	        & 1.95 & & 96.67      & 0.66 \\
JRC(q=p=2)  & 99.33	    & 0.97 & & 98.67      & 0.17 \\
SVM        & 96.67	    & 5.09 & & 96.67      & 1.46 \\
JRC(q=1.5,p=1) & 99.33  & 4.89 & & 98.67      & 3.86 \\
JRC(q=1.5,p=0.5)& 99.33 & 4.89 & & 98.67      & 3.89 \\
JRC(q=p=1)     &  99.33 & 5.54 & & 99.33      & 1.11 \\
JRC(q=1,p=0.5) &  99.33 & 4.79 & & 99.33      & 1.09 \\
\hline
\end{tabular}
\caption{\label{tab2}The recognition accuracy (\%) and CPU time (second)\protect \\ for Georgia-Tech database}
\end{table*}

\begin{table*}[!htb]
\centering
 \begin{tabular}{lccccc}
 \hline
 & \multicolumn{2}{c}{Down sampling ratio 1/8}& &\multicolumn{2}{c}{Down sampling ratio 1/16}\\
\hline
Methods  &  Accuracy    & Time & &  Accuracy  & Time \\ \hline\hline	
SRC      &  96.76	    & 4828 & &	96.36     & 668.53 \\
JRC(q=2,p=1) & 96.96	& 22.67 & &  76.11     & 164.71 \\
CRC-RLS & 96.76	        & 2.02 & & 95.55      & 1.9 \\
JRC(q=p=2)  & 96.96	    & 0.75 & & 91.29      & 0.34 \\
SVM        & 95.55	    & 6.12 & & 94.33      & 2.61 \\
JRC(q=1.5,p=1) & 96.96  & 22.04 & & 87.05      & 22.03 \\
JRC(q=1.5,p=0.5)& 96.96 & 54.21 & & 65.59      & 101.59 \\
JRC(q=p=1)     &  96.96 & 27.08 & & 90.49      & 20.51 \\
JRC(q=1,p=0.5) &  96.96 & 26.87 & & 91.29      & 25.23 \\
\hline
\end{tabular}
\caption{\label{tab3}The recognition accuracy (\%) and CPU time (second)\protect \\ for Extended Yale B database}
\end{table*}

Based on the experimental results on three databases,  we draw the following conclusions:

\noindent$\bullet$ Jointly representing all the testing images simultaneously does accelerate  face recognition. On all the databases, JRC (q=p=2) is the fastest one. The CPU time is thousand times less than that of SRC. For example, JRC (q=p=2) classifies $484$ images in $0.17$ second on Georgia-Tech database with downsampling ratio $1/16$. And the accuracy rate is $98.67\%$, outperforming SRC ($97.33\%$), CRC-RLS ($96.67\%$) and SVM ($96.67\%$). More details can be found in Table \ref{tab1}-\ref{tab3}.

\noindent$\bullet$ JRC exhibits competitive performance in recognition accuracy. On AT \& T database, the recognition rate of JRC is 97.5\%, compared to 98.75\% for SRC, 95\% for CRC-RLS and SVM. On Georgia-Tech database, JRC achieves the best recognition rate (99.33\%), consistently exceeds other classification schemes. On Yale B database with downsampling ratio $1/8$, JRC also outperforms other methods in accuracy. Unfortunately, JRC does not keep the best achievement on downsampling ratio $1/16$. The possible reason is that some pictures with strong contrast of lighting (see Figure \ref{fig3}) aggravates the noise for other images in joint coding.

\begin{figure}[!htb]
\centering
\subfigure{\includegraphics[width=40mm]{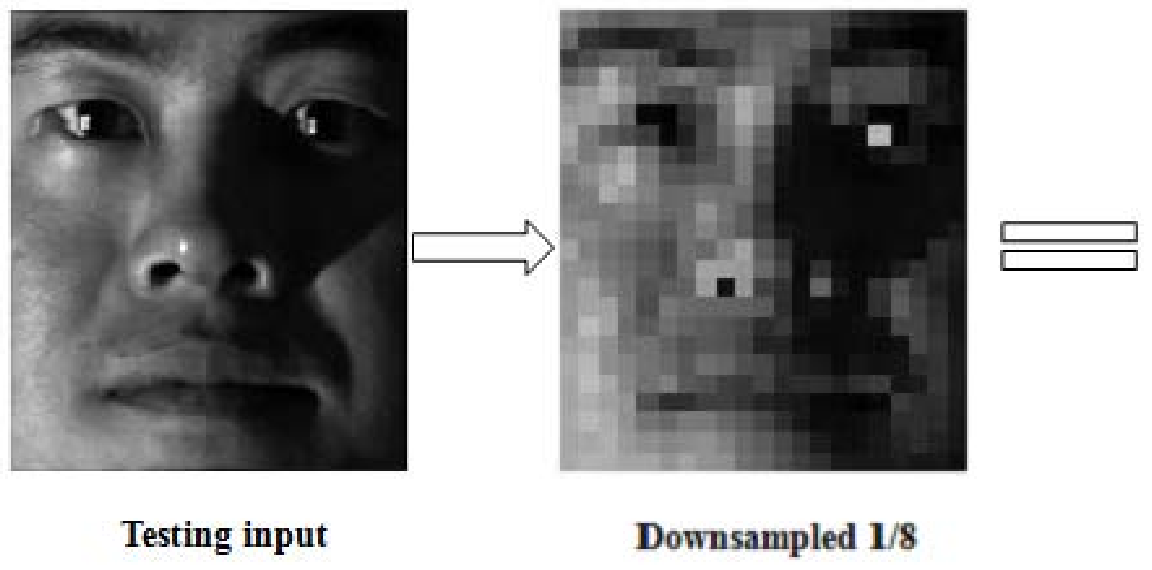}}
\subfigure{\includegraphics[width=38mm]{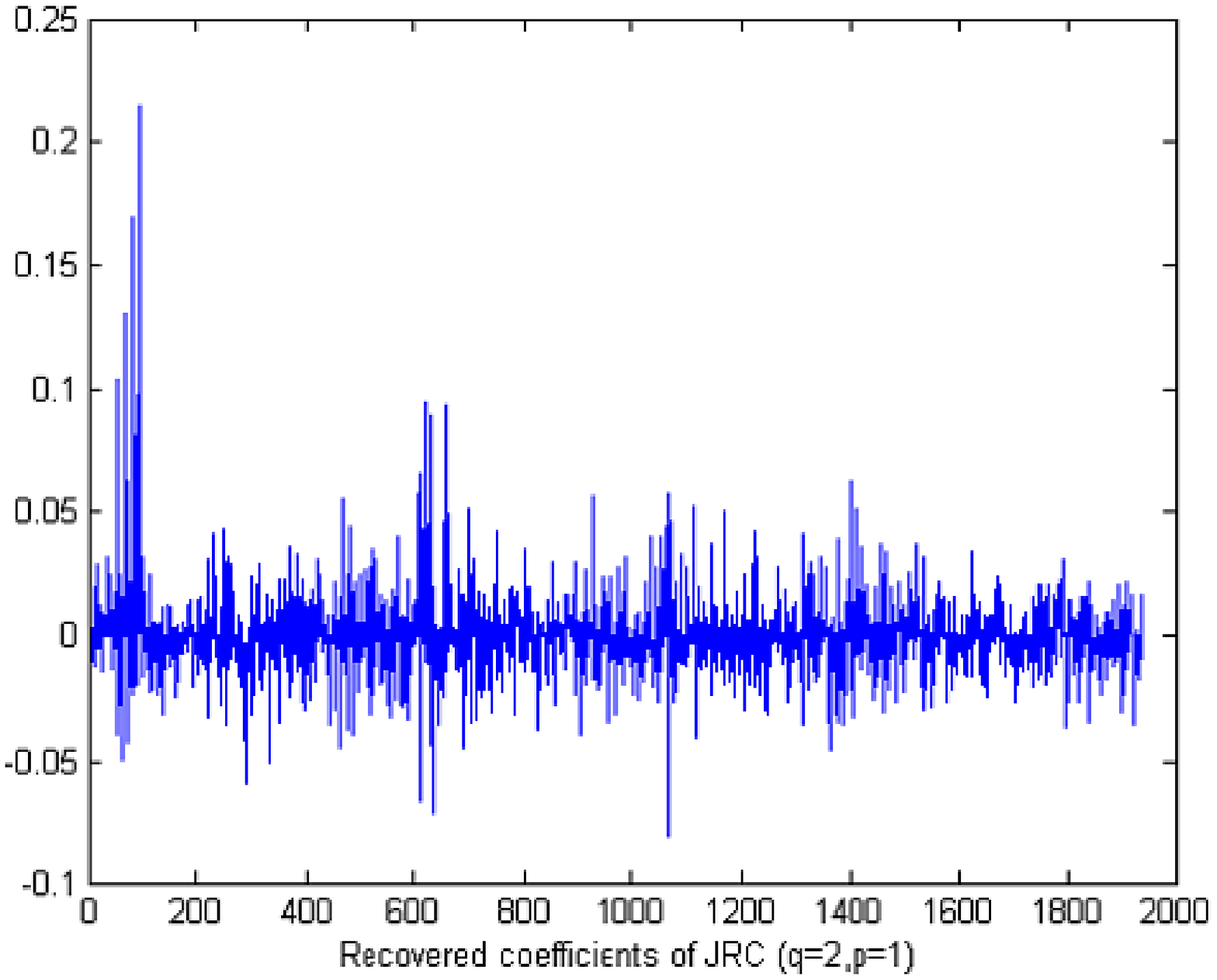}}
\subfigure{\includegraphics[width=38mm]{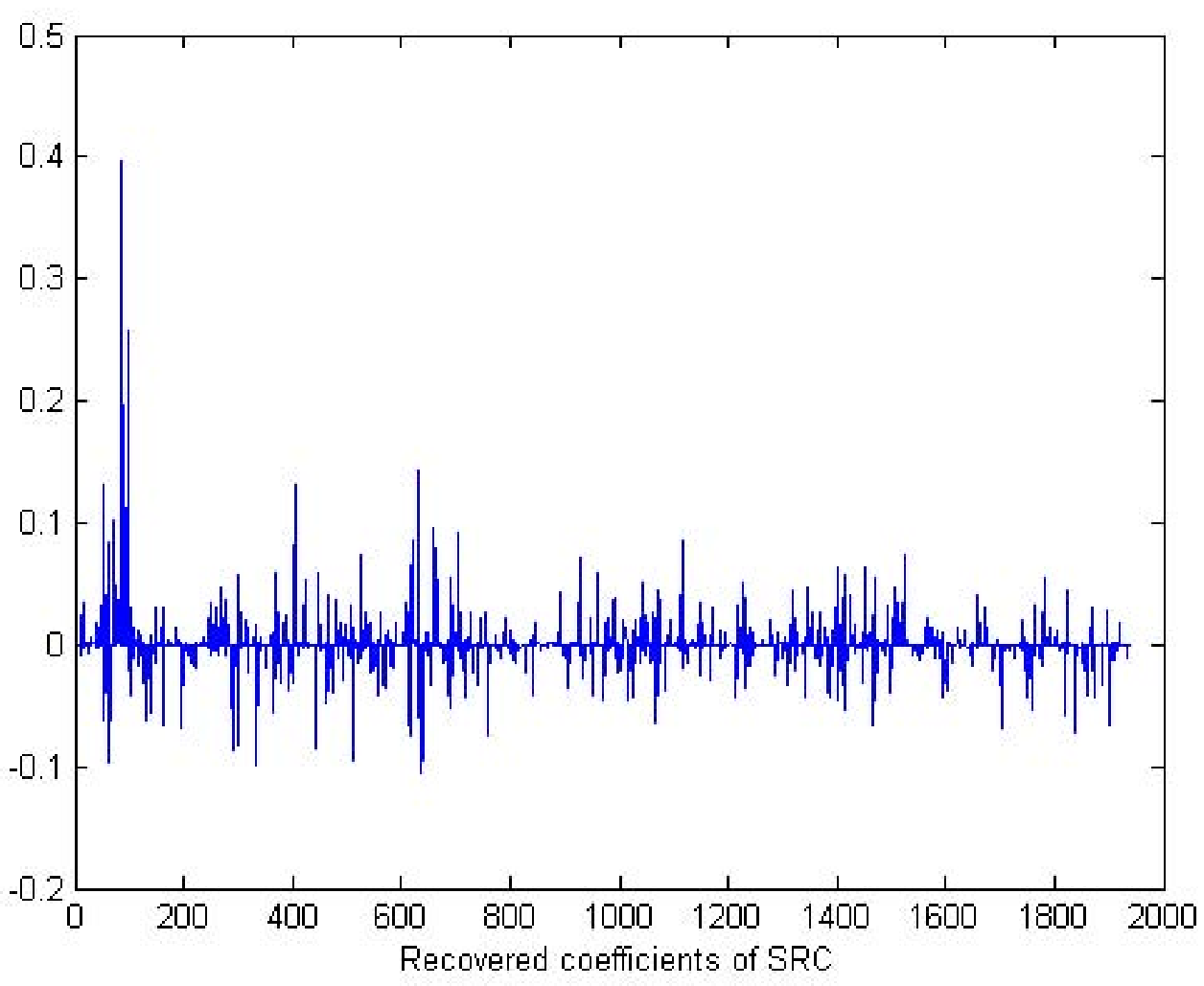}}
\caption{\label{fig4} The recovered coefficients by JRC (q=2,p=1) and SRC}
\end{figure}

\noindent$\bullet$ Different $q\in [1, 2]$ and $p\in (0, 2]$ for JRC indicate different feature pattern behind in the image set. Taking JRC ($q=2,p=1$) for example, the joint model combines sparsity of representation and correlation of multiple images. The representation coefficients reveal the joint effect on JRC ($q=2,p=1$), Figure \ref{fig4} gives an example from Yale B database. Compared to SRC, JRC ($q=2,p=1$) concentrates a group sparsity but not a single one. Acutally, the other testing samples (12 pictures) of the same subject also have the similar group representation pattern.

\begin{figure}[!htb]
\centering
\subfigure[]{\includegraphics[width=38mm]{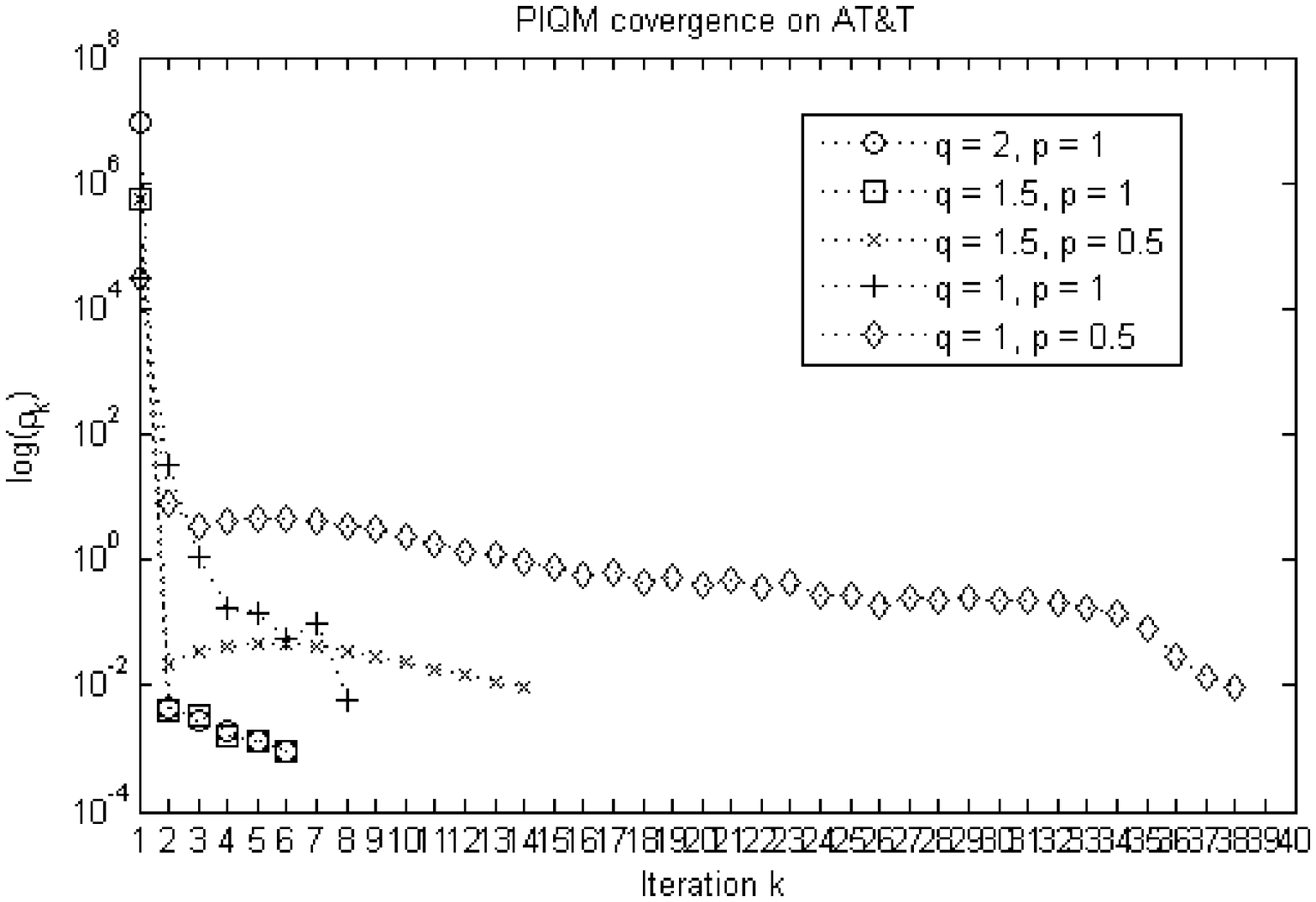}}
\subfigure[]{\includegraphics[width=38mm]{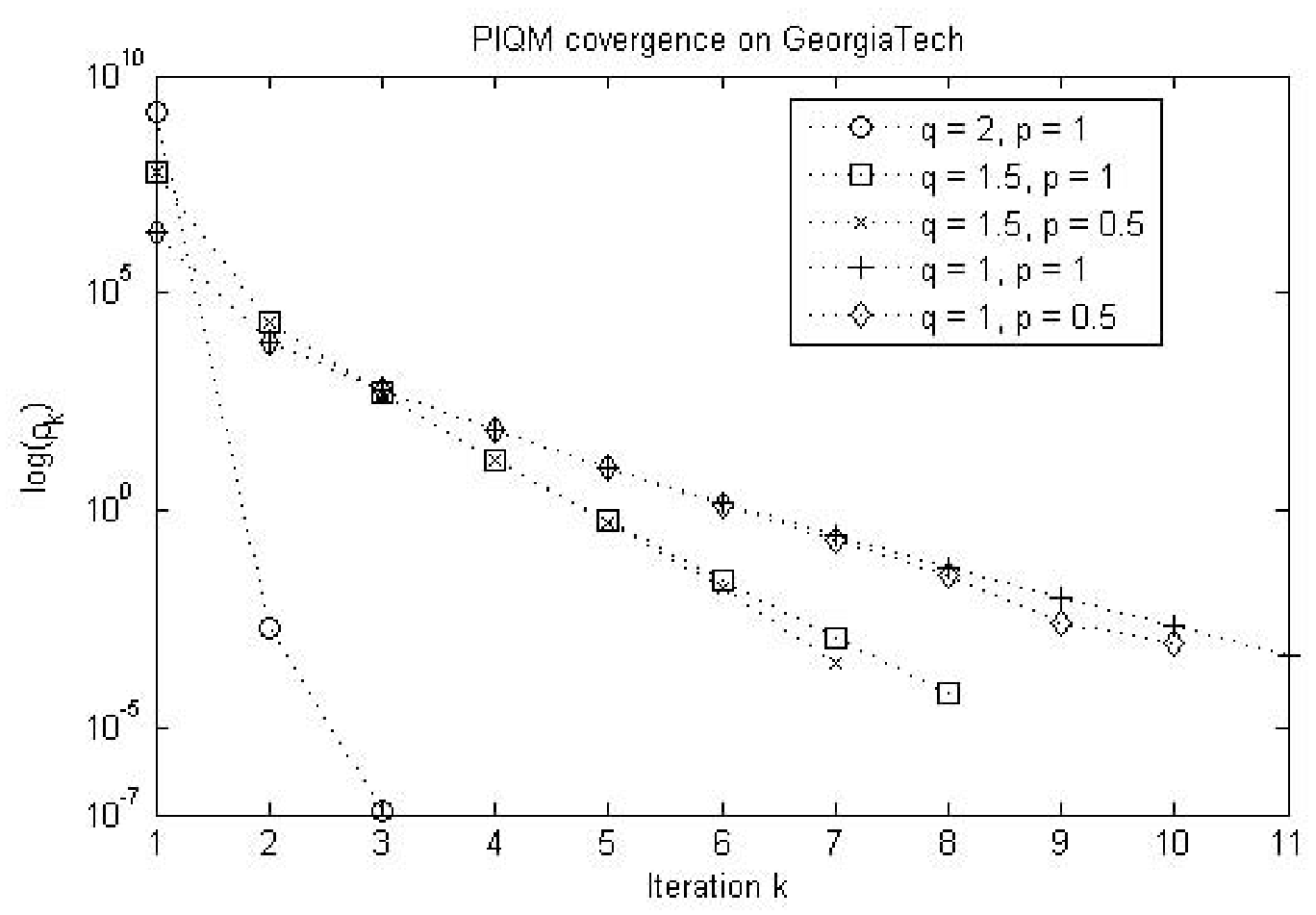}}
\subfigure[]{\includegraphics[width=38mm]{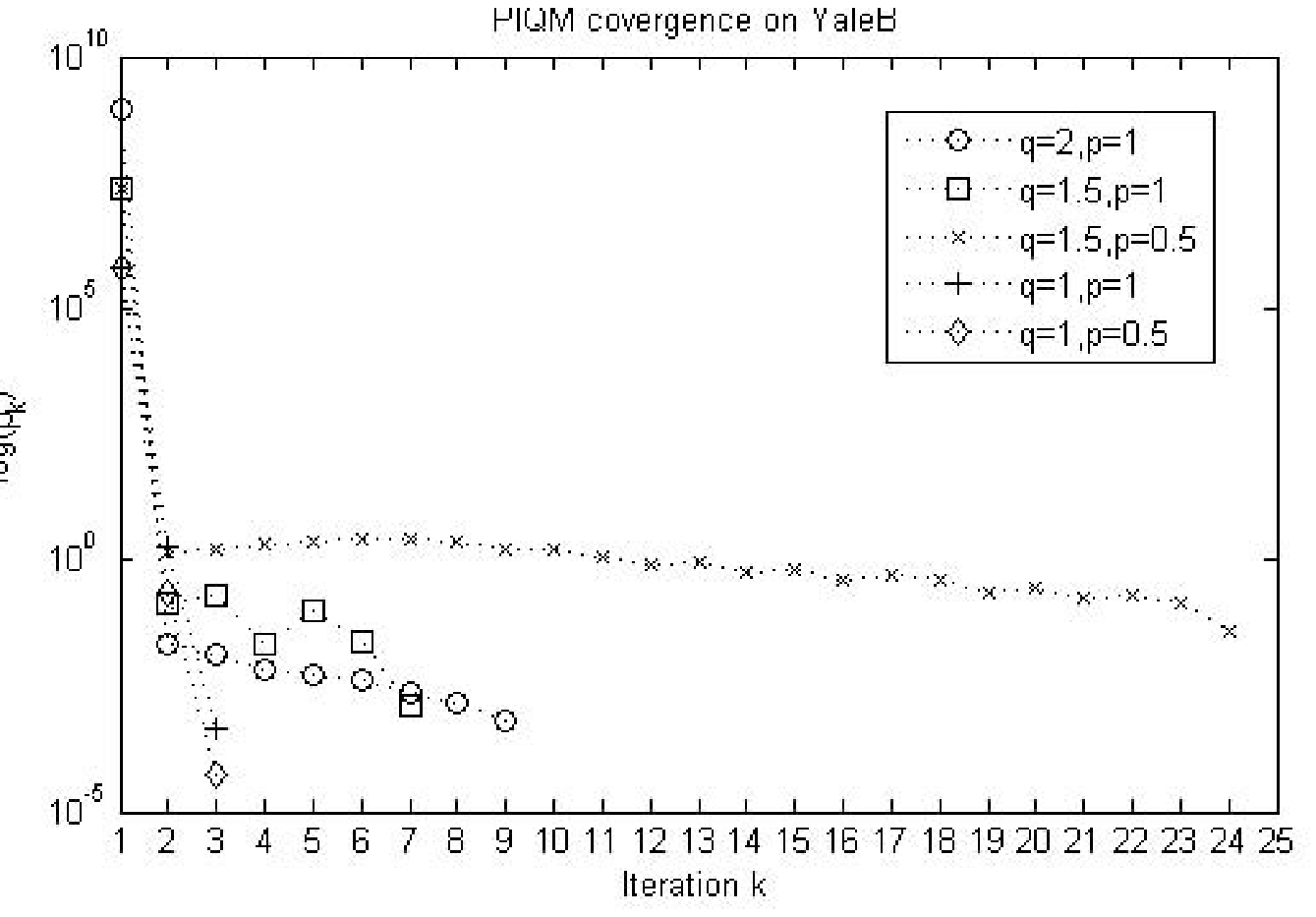}}
\caption{\label{fig5} (a) PIQM on AT \& T(b) PIQM on Georgia-Tech (c) PIQM on Yale B}
\end{figure}

\noindent$\bullet$ The convergence behavior of PIQM for JRC is displayed in Figures \ref{fig5}. The x axis is the iterations and y-axis stands for the logarithm of $\rho_k$. PIQM converges within $40$ steps on three databases for all jointly sparse models (five pairs $q$ and $p$). JRC (q=p=2) always converges in three iterations hence its plot is omitted here. Anyway, PIQM provides a uniform algorithm for varied JRC with respect to $q\in [1,2]$ and $p\in (0,2]$.

\noindent$\bullet$ From Table \ref{tab1}-\ref{tab3}, it is observed that CRC-RLS has a fairly good performance in recognition accuracy and CPU time. But CRC-RLS is heavily sensitive to the regularization parameter $\lambda$ (see Table \ref{tab4}) because it has a smooth regularizer. By comparison, JRC (q=p=2) is more stable for its joint technique. Multiple images has complementary effect for recognition especially when the model is ill-posed.
\begin{table*}[!htb]
\centering
\begin{tabular}{lccccc}
\hline
$\lambda=$  &  0.01    & 0.1 &1 &  10  & 100 \\ \hline\hline	
CRC-RLS & 28.34	        & 66.82 & 95 & 96.76      & 96.76 \\
JRC(q=p=2)  & 96.96	    & 96.96 &96.96 & 96.96      & 96.96 \\
\hline
\end{tabular}
\caption{\label{tab4}The recognition accuracy (\%) for different $\lambda$  on Extended Yale B database with downsampling ratio $1/8$}
\end{table*}

\section{Conclusions}
In this paper, a joint representation classification  for collective face recognition is proposed. By aligning all the testing images into a matrix, joint representation coding is reduced to a kind of generalized matrix pseudo norm based optimization problems. A unified algorithm is developed to solve the mixed  $l_{2,q}-l_{2,p}$-minimizations for $q\in [1,2]$ and $p\in (0, 2]$.  The convergence is also uniformly demonstrated. To adapt the algorithm to the large scale case, a practical iterative quadratic method is considered to inexactly solve the subproblems. Experiment results on three data-sets validate the collective performance of the proposed scheme. The joint representation based classification is confirmed to improve the performance in recognition rate and running time than the state-of-the-arts.

\vskip 1cm

\noindent{\bf\large Acknowledgement} The first author thanks software engineer Luo Aiwen for his code support.

\end{document}